\pdfoutput=1
\documentclass[10pt,twocolumn,letterpaper]{article}

\usepackage{iccv}
\usepackage{times}
\usepackage{epsfig}
\usepackage{graphicx}
\usepackage{amsmath}
\usepackage{amssymb}

\usepackage{booktabs}
\usepackage{color}
\usepackage{enumitem}
\usepackage{mathtools}
\usepackage{subfigure}
\usepackage{makecell}
\usepackage{soul}

\newcommand{\losssparse}{\mathcal{L}_{\textnormal{sparsity}}}
\newcommand{\losstemp}{\mathcal{L}_{\textnormal{temporal}}}
\newcommand{\losssaliency}{\mathcal{L}_{\textnormal{saliency}}}

\DeclareGraphicsExtensions{.png,.pdf,.eps,.jpg}

\def\figurePath{figs/}
\def\myfigure#1#2{\begin{figure}[t!]\centering\includegraphics*[width = \linewidth]{\figurePath#1}\caption{#2}\label{fig:#1}\end{figure}}

\def\mycfigurew#1#2#3{\begin{figure*}[t]\centering\includegraphics*[clip, width = #3\linewidth]{\figurePath#1}\caption{#2}\label{fig:#1}\end{figure*}}

\newcommand{\refSec}[1]{Sec.~\ref{#1}}

\newcommand{\refFig}[1]{Fig.~\ref{fig:#1}}

\newcommand{\mycomment}[1]{}

\definecolor{darkred}{rgb}{0.6,0,0}
\definecolor{green}{rgb}{0.0,0.5,0}
\definecolor{blue}{rgb}{0,0,0.75}
\definecolor{orange}{rgb}{1,0.6,0.2}
\definecolor{red}{rgb}{1,0,0}

\soulregister\ref{7}
\soulregister\cite{7}
\soulregister\refFig{7}

\usepackage[table,x11names]{xcolor}
\usepackage{stfloats}

\usepackage[accsupp]{axessibility}  

\usepackage[pagebackref=true,breaklinks=true,letterpaper=true,colorlinks,bookmarks=false]{hyperref}

\iccvfinalcopy 

\def\iccvPaperID{7964} 

\ificcvfinal\fi

\begin{document}

\title{The Way to my Heart is through Contrastive Learning:\\Remote Photoplethysmography from Unlabelled Video}

\author{{John Gideon\thanks{Equal contribution} \; \; \; Simon Stent\footnotemark[1]}\\
Toyota Research Institute\\
Cambridge, MA, USA\\
{\tt\small \{john.gideon, simon.stent\}@tri.global}
}

\maketitle
\ificcvfinal\thispagestyle{empty}\fi



\begin{abstract}
The ability to reliably estimate physiological signals from video is a powerful tool in low-cost, pre-clinical health monitoring. In this work we propose a new approach to remote photoplethysmography (rPPG) -- the measurement of blood volume changes from observations of a person's face or skin. Similar to current state-of-the-art methods for rPPG, we apply neural networks to learn deep representations with invariance to nuisance image variation. In contrast to such methods, we employ a fully self-supervised training approach, which has no reliance on expensive ground truth physiological training data. Our proposed method uses contrastive learning with a weak prior over the frequency and temporal smoothness of the target signal of interest. We evaluate our approach on four rPPG datasets, showing that comparable or better results can be achieved compared to recent supervised deep learning methods but without using any annotation. In addition, we incorporate a learned saliency resampling module into both our unsupervised approach and supervised baseline. We show that by allowing the model to learn where to sample the input image, we can reduce the need for hand-engineered features while providing some interpretability into the model's behavior and possible failure modes.
We release code for our complete training and evaluation pipeline to encourage reproducible progress in this exciting new direction.\footnote{\url{https://github.com/ToyotaResearchInstitute/RemotePPG}}
\end{abstract}

\section{Introduction}
\label{sec:intro}

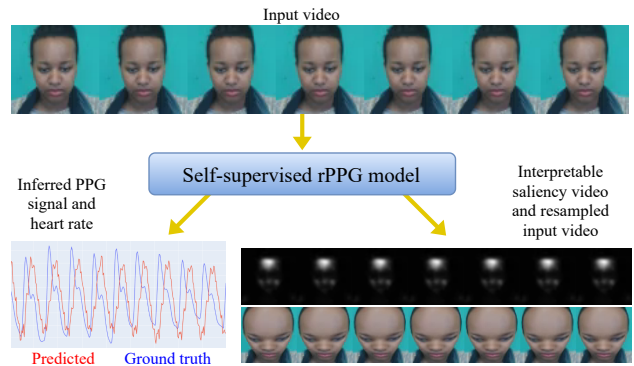
\myfigure{Fig1_contrastive}{From a video of a person's face alone, our model learns to estimate the person's cardiac activity in the form of a photoplethysmographic (PPG) signal (left) observed through temporal patterns in the video, as well as a saliency signal (right) which shows where in the video the model's estimated activity is strongest (in this case, the center of the forehead). We show through extensive experiments on video datasets with physiological ground truth that our approach can match and sometimes even improve upon existing end-to-end supervised methods, while providing both \textit{interpretability} into the model behavior and incurring \textit{zero annotation cost} to train. The figure above shows a real output from our model trained on the UBFC~\cite{bobbia2019unsupervised} dataset. We note that the phase offset between predicted and ground truth PPG signals may be due to synchronization issues between the video and ground truth itself - a detail discussed further in~\refSec{sec:method:loss}.}

Understanding the physiological state of a person is important in many application areas, from health and fitness through to human resource management and human machine interaction.
Conventional approaches to estimate such information, such as electrocardiograms (ECG) or photoplethysmograms (PPG), require interaction with the subject and are troublesome to setup, limiting their usefulness and scalability.
In recent years, research utilizing advances from the field of computer vision and machine learning has explored and improved upon methods for passively monitoring physiological information from videos of subjects.

In this work we introduce a new method for remote photoplethysmography (rPPG), or imaging PPG, a technique in which changes in transmitted or reflected light from the body due to volumetric changes in blood flow are measured at a distance using a standard imaging device. This differs from the more intrusive contact PPG, in which the same signal is measured at peripheral body tissues such as the fingertips via a contact sensor which projects and measures reflected LED light. Compared to PPG, the signals for remote PPG are often too subtle for the human eye to perceive, but under certain illumination conditions, can be isolated and magnified in digital imagery if one knows where to look~\cite{Wu12Eulerian}.
By finding these signals and using them to estimate underlying cardiac activity, particularly from webcam-quality video such as shown in the input video of~\refFig{Fig1_contrastive}, rPPG can therefore help to meet a need for low-cost, non-contact health monitoring.

In the field of computer vision, many researchers have tackled the problem of rPPG in the past, leaning on a wide variety of techniques from signal processing and machine learning (see e.g.~\cite{chen_deepphys_2018,li_remote_2014,niu_video-based_2020,tulyakov2016self,yu2019remote}).
Recent efforts have tended to favor deep learning, which is known for solving particular tasks well by discovering feature representations that are robust to many forms of nuisance variation.
In rPPG, such variation takes the form of lighting changes, motion, and changes in facial appearance or gesture, all of which can easily obscure the underlying PPG signal.
Supervised deep learning approaches to rPPG such as~\cite{chen_deepphys_2018,lee_meta-rppg_2020,niu_video-based_2020,yu2019remote} have shown that, with annotated data to train on, rPPG can be achieved with higher robustness to such variation.
However, the cost of annotated data is not cheap, due to the need to equip subjects with contact PPG or ECG sensors while capturing data. It is therefore hard to scale the capture of such datasets, although, driven by data-hungry algorithms, there have been recent efforts in this direction~\cite{niu_accv_2018_viplhr,niu_video-based_2020}.

In this work we take a contrary approach to applying deep learning to rPPG. We view the problem through the lens of self-supervised learning, and in doing so bridge the data economy of older approaches with the robustness of learned representations.
Our contrastive training approach is built around three assumptions about the underlying signal of interest:
\begin{enumerate}[label=A\arabic*]
    \item We assume the signal of interest lies within a certain range. We set this range for the rest of the paper at 40 to 250 beats per minute, which captures the vast majority of human heart rates~\cite{AmericanHeartAssociation}.
    \item We assume the signal of interest typically does not vary rapidly over short time intervals: the heart rate of a person at time $t$ is similar to their heart rate at $t+W$, where W is in the order of seconds.
    \item Finally, the signal has some visible manifestation (even if undetectable to the human eye) and is the dominant visual signal within the target frequency range.
\end{enumerate}

\textbf{Contributions.} We show that by setting up a \textbf{contrastive learning} framework based on these assumptions, it is possible to train a deep neural network to estimate the PPG signal (and therefore track the heart rate) of a subject from video of their face, entirely without ground truth training data. We introduce novel loss functions for both supervised and contrastive training that are robust to de-synchronization in the ground truth and take advantage of our above assumptions. Moreover, since the behavior of a deep neural network regressing PPG alone may be difficult to comprehend or have confidence in without access to annotated data, we propose a front-end \textbf{saliency-based sampling} module, inspired by~\cite{recasens2018learning}, to accentuate the parts of the input data which are most relevant to a back-end PPG estimator. A by-product of this, as shown in~\refFig{Fig1_contrastive}, is that our model can also output interpretable saliency maps. These maps provides some transparency into the spatial location of the model's discovered signal of interest; in this case, the subject's forehead and parts of her nose and cheeks, which matches the conventional understanding of where the rPPG signal is strongest~\cite{irani2014improved,moreno2015facial}. We note that these contributions are independent but complementary. As shown in our experiments, the saliency sampler can be appended to both supervised and contrastive models with similar effect, and our contrastive model can learn to predict PPG without the saliency sampler.
Finally, we unify existing freely available PPG video datasets and provide our complete training and evaluation pipeline to encourage further reproducible efforts in what we believe is an exciting direction of research in computer vision for human health monitoring.

\section{Background and Related Work}
\label{sec:related}

\textbf{Remote photoplethysmography.} 
Approaches to rPPG, or heart rate (HR) estimation from video, can typically be broken down into three components: (i) a pre-processing stage, to minimize nuisance variation (for example through face detection and tracking) and discard irrelevant information in the input data; (ii) a PPG signal extraction stage; and (iii) a heart rate estimation stage from the estimated PPG signal.
Early work in rPPG focused on finding signals within the image which were more easily accessible and perhaps more robust to nuisance variation, such as color over specific regions~\cite{kwon2012validation,poh2010advancements,poh2010non} or motion~\cite{balakrishnan2013detecting}. As dense facial tracking improved, the pre-processing part of the pipeline increased in complexity, incorporating techniques like landmark detection~\cite{li_remote_2014,tulyakov2016self}, skin segmentation~\cite{bobbia2019unsupervised,fouad2019optimizing} and carefully engineered ROI-based feature extraction~\cite{lee_meta-rppg_2020,niu_video-based_2020}.
Of these, our work is most similar in spirit to~\cite{tulyakov2016self}, who describe an approach based on self-adaptive matrix completion to simultaneously estimate the heart rate signal and (learn to) select reliable face regions at each time. However, unlike their method, which requires keypoint tracking and careful image warping, ours is not necessarily face-specific and is arguably less aggressive in discarding potentially useful information: we do not reduce our feature space to chrominance features but instead pass the raw, warped image data to our PPG estimator.

In recent years, the reliance on relatively clean and stable input data has been slowly lifted as methods based on deep learning have increasingly proved themselves capable of learning more robustly through noise~\cite{chen_deepphys_2018,lee_meta-rppg_2020,niu_accv_2018_viplhr,niu_video-based_2020,yu_remote_2019_physnet,yu2019remote,spetlik_visual_2018}.
HR-CNN~\cite{spetlik_visual_2018} uses a two-stage convolutional neural network (CNN) with a per-frame feature extractor and an estimator network. DeepPhys~\cite{chen_deepphys_2018} uses a VGG-style CNN with separate predictions branches tailored towards motion and intensity. PhysNet~\cite{yu_remote_2019_physnet} investigates both a 3DCNN based model and a model that combines 2DCNN and LSTM to learn spatio-temporal features. Yu~\etal address the issue of rPPG detection in highly compressed facial videos by adding an additional autoencoder video enhancement stage to their model \cite{yu2019remote}.
More recently, state of the art performance in rPPG has been achieved by a cross-verified feature disentangling strategy~\cite{niu_video-based_2020}, which helps to isolate information which is most pertinent to physiological signal estimation. Their method involves computing hand-designed facial features called MSTmaps, which are average-pooled RGBYUV values across various combinations of regions of interest on the face. Their multi-branch output can be used to estimate both heart rate and PPG signal (as well as other possible signals such as respiratory frequency), with the joint loss from estimating both simultaneously returning further performance improvements.
Finally, the RepNet model~\cite{Dwibedi_2020_CVPR} demonstrated an effective way to estimate the period of repeated actions in video by computing the self-similarity of image representations. The authors show that, when applied to stable facial video preprocessed by~\cite{Wu12Eulerian}, their model can recover human pulse.

As pointed out in the recent meta-RPPG work of Lee~\etal~\cite{lee_meta-rppg_2020}, significant changes often occur in data distributions between model training and deployment, which can detrimentally affect the real-world performance of otherwise state-of-the-art end-to-end supervised learning approaches. To cope with such shifts, the authors propose a transductive meta-learner which can perform self-supervised weight adjustment from unlabelled samples. Our model is similar in intent, but rather than relying on modelling the domain shift, we allow for self-supervised training within entirely new domains from scratch. Crucially, since our approach still relies on an ``end-to-end'' trained deep neural network, it may be favorable compared to more traditional methods which require significant feature engineering or signal processing, since it is likely to improve with data.

\textbf{Contrastive learning.}
To learn richer feature representations of data, contrastive learning~\cite{chen2020simple,grill2020bootstrap,Hadsell06dimreduction,he2020momentum} proposes augmenting the data with different versions of itself during training, and contrasting a model's representations of the data in ways that encourage learning features with invariance or equivariance to particular augmentations. 
For example, in~SimCLR~\cite{chen2020simple}, augmentations include spatial distortion (cropping, rotating, blurring) and chromatic distortion, and features can be learned which can help identify the semantic identity of objects in images with less sensitivity to such distortions.
Unlike these techniques, in our case the signal of interest is stronger temporally than it is in an individual image, where it remains mostly imperceptible to the human eye. We therefore deliberately avoid image-domain augmentation, and focus instead on frequency augmentation. Specifically, by resampling video sequences at different rates and forcing the network to learn to detect whether two videos have similar or dissimilar underlying signals, we show that it is possible for the network to learn to filter particular signals of interest from the input data.

In the context of rPPG, one issue with using contrastive learning may be that, in the absence of annotated data to evaluate with, it is hard to understand model performance. 
To allow for some transparency into the model's inner workings, we equip it with a saliency sampling layer which shows which particular parts of the image were used by the system. Our approach builds on prior work~\cite{recasens2018learning}, but applies the sampling layer to a self-supervised task and introduces additional priors to improve the behavior of the saliency map. While strictly not necessary for our approach to work, the sampler provides the system with an interpretable intermediate output, which can allow a practitioner to determine whether or not the network has converged to a sensible solution when training without ground truth data.

\section{Method}
\label{sec:method}

An overview of our approach is shown in~\refFig{system}. We now describe each stage of the pipeline in sequence.

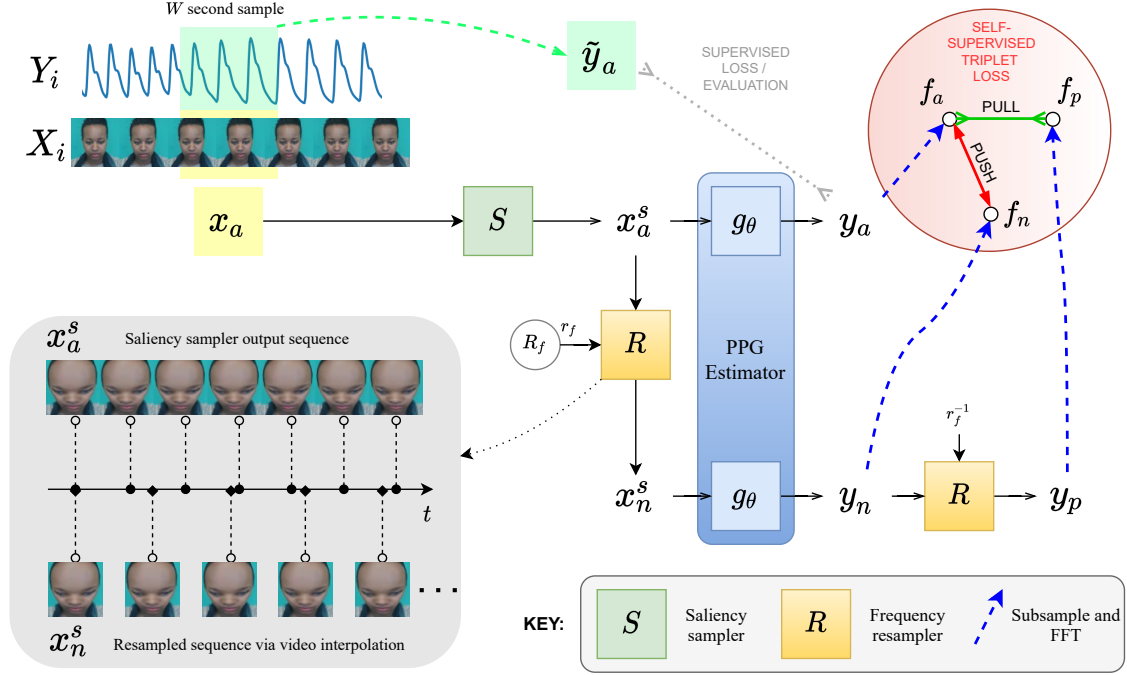
\mycfigurew{system}{\textbf{Overview of our approach.} We first sample a video clip, $x_a$, of length $W$ from the source video. This video is passed through the saliency sampler, $S$, to generate the warped anchor, $x_a^s$. The anchor is passed through a PPG Estimator $g_\theta$ to get $y_a$. If supervised training is employed, we employ a maximum cross-correlation (MCC) loss between the ground truth $\tilde{y}_a$ and $y_a$. If instead contrastive training is used, a random frequency ratio $r_f$ is sampled from a prior distribution. The warped clip $x_a^s$ is then passed through the frequency resampler $R$ to produce the negative sample $x_n^s$, showing a subject with an artificially higher heart rate. This sample is passed through $g_\theta$ to produce the negative example PPG $y_a$. The negative sample is again resampled with the inverse of $r_f$ to produce a positive example PPG $y_p$. Finally, the contrastive loss, MVTL, is applied to the PPG samples, using a PSE MSE distance metric. For further details about metrics and losses used, please see~\refSec{sec:method:loss}.}{0.87}
\subsection{Preprocessing}
\label{sec:method:preprocess}

For all datasets studied, we adopt a simple preprocessing procedure. We first estimate a bounding box around the face~\cite{zhang2017s3fd} and add an additional 50\% scale buffer to the box before extracting a $192\times128$ frame. We only update the buffered bounding box location on subsequent frames if the new non-buffered bounding box is outside the larger one. This ensures relative stability of the video, while still allowing for occasional movement and re-alignment. Because the datasets used in this paper do not include full body or camera motion, re-alignment occurs infrequently. We finally crop and scale the videos to a $64\times64$ resolution for input to the model, which we found to return comparable performance to larger resolutions at lower computational cost.

\subsection{Saliency Sampler}
\label{sec:method:saliency}
Input video sequences are passed to an optional saliency sampler module, building on work from~\cite{recasens2018learning}. In the context of our model, the module's purposes is two-fold: firstly, to provide transparency as to what the PPG estimator is learning, which is particularly valuable if little annotated data exists to validate with; and secondly, to warp the input image to spatially emphasize task-salient regions, before passing it on to the task network (the PPG estimator) as described in~\cite{recasens2018learning}. In all experiments we use a pre-trained ResNet-18~\cite{he2016deep}, truncated after the \texttt{conv2\_x} block, which we empirically find to perform well for the task without incurring significant computational overhead. Diverging from~\cite{recasens2018learning}, we optionally impose two additional loss terms: 
\begin{eqnarray}
 \losssparse &=& - \frac{1}{ND} \sum_{i}^D \sum_{j}^{N} s_i^j \log (s_i^j) \\
 \losstemp &=& \frac{1}{N(D-1)} \sum_i^{D-1} \sum_{j}^N (d_{i, i+1}^j)^2 \\
 \losssaliency &=& w_s \losssparse + w_t \losstemp
\end{eqnarray}
where $s_i^j$ is the value of the per-frame softmax-normalized saliency map at frame $i$ in the sequence and position $j$ in the frame, $d_{i,i+1}^j$ is the difference between saliency pixel $j$ from frame $i$ to $i+1$, $N$ is the number of pixels in a frame and $D$ is the number of frames in the video sequence.
The sparsity term favors solutions which have lower entropy (i.e. are spatially sparse, such as the forehead in~\refFig{Fig1_contrastive}),  while the temporal consistency favors solutions that are smooth from frame to frame.

\begin{table*}
\begin{small}
\begin{center}
\begin{tabular}{lll}
\toprule
Loss function & Assumptions & Used by \\
\midrule
Negative max cross correlation (MCC) & HR is within a known frequency band (Assumption 2) & Supervised loss \\
Multi-view triplet loss (MVTL) & HR is stable within a certain window (Assumption 1) & Contrastive loss \\ 
Power spectral density mean squared error (PSD MSE) & HR is within a known frequency band (Assumption 2) & Distance metric \\
Irrelevant power ratio (IPR) & HR is within a known frequency band (Assumption 2) & Validation metric \\
\bottomrule
\end{tabular}
\end{center}
\caption{\textbf{Loss functions and metrics.} The losses used during training, the distance metric used for contrastive self-supervision, and the validation metric used during self-supervision. All supervised losses are also used as supervised validation metrics.}
\label{table:losses}
\end{small}
\end{table*}

\subsection{PPG Estimator}
\label{sec:method:models}

We use a modified version of the 3DCNN-based PhysNet architecture as our PPG estimator \cite{yu_remote_2019_physnet}. The core of PhysNet is a series of eight 3D convolutions with a kernel of (3, 3, 3), 64 channels, and ELU activation. This allows for the network to learn spatio-temporal features over the input video. Average pooling and batch normalization are also employed between layers. In the PhysNet paper, two transposed convolutions are used to return the encoded representation to the original length. However, we found that these introduced aliasing in the output PPG signal. We modify this part of the network to instead use upsampling interpolation ($\times4$) and a 3D convolution with a (3, 1, 1) kernel. This upsampling step is repeated twice and removes the aliasing. Empirically, we found it to improve test RMSE by 0.2 bpm on average across datasets. Next, we perform adaptive average pooling to collapse the spatial dimension and produce a 1D signal. A final 1D convolution is applied to convert the 64 channels to the output single channel PPG. Please see the supplementary material for full architectural details.

\subsection{Loss Functions}
\label{sec:method:loss}

We use a variety of loss functions and metrics during training, as summarized in Table~\ref{table:losses}. 

We propose \textbf{maximum cross-correlation} (MCC) as a new loss function and metric for rPPG supervised training. While PC assumes PPG synchronization, MCC determines the correlation at the ideal offset between signals. This causes the loss to be more robust to random temporal offsets in the ground truth, assuming heart rate is relatively stable (Assumption 1). The authors of meta-RPPG \cite{lee_meta-rppg_2020} adopt a similar approach with their use of an ordinal loss. However, this requires the model to learn an ordinal regression instead of a raw PPG signal. MCC can be computed efficiently in the frequency domain, as follows:
\begin{equation}
\textnormal{MCC} = c_{pr} \times \textnormal{Max} \left( \frac{F^{-1}\{\textnormal{BPass}(F\{y\} \cdot \overline{F\{\hat{y}\}})\}}{\sigma_y \times \sigma_{\hat{y}}} \right)
\end{equation}
We first subtract the means from each signal to simplify the calculation - resulting in $y$ and $\hat{y}$. Cross-correlation is then calculated in the frequency domain by taking the fast Fourier transform (FFT) of the two signals and multiplying one with the conjugate of the other. To prevent circular correlation, we zero-pad the inputs to the FFT to twice their length. We apply a band bass filter by zeroing out all frequencies outside the range of expected heart rate (40 to 250 bpm), enforcing our Assumption 2. We then take the inverse FFT of the filtered spectrum and divide by the standard deviation of the original signals, $\sigma_y$ and $\sigma_{\hat{y}}$, to get the cross-correlation. The maximum of this output is the correlation at the ideal offset. We scale the MCC by a constant, $c_{pr}$, which is the ratio of the power inside the heart rate frequencies. This ensures the MCC is unaffected by the frequencies outside the relevant range. We use MCC as our loss function for supervised training. In the supplementary material we include an analysis of the robustness of MCC to randomly injected ground truth synchronization error, showing its more stable performance compared to more standard losses such as Pearson's correlation and the signal-to-noise ratio.

We also introduce \textbf{multi-view triplet loss} (MVTL) as our loss function for contrastive training. As shown in~\refFig{system}, our self-supervised pipeline has three output branches - anchor ($y_a$), positive ($y_p$), and negative ($y_n$). From these three branches, we take $V_{N}$ subset views of length $V_{L}$. This enforces Assumption 1 - that heart rate is relatively stable within a certain window. Because of this, the signal within each view should appear similar. We then calculate the distance between all combinations of anchor and positive views ($P_{tot}$) and all combination of anchor and negative views ($N_{tot}$). We calculate $P_{tot} - N_{tot}$ and scale by the total number of views, $V_{N}^2$, to get the final loss.

We use the \textbf{power spectral density mean squared error} (PSD MSE) as the distance metric between two PPG signals when performing contrastive training with MVTL. We first calculate the PSD for each signal and zero out all frequencies outside the relevant heart rate range from 40 to 250 bpm (Assumption 1). We then normalize each to have a sum of one and compute the MSE between them.

Finally, we use the \textbf{irrelevant power ratio} (IPR) as a validation metric during contrastive training. We first calculate the PSD and split it into the relevant (40 to 250 bpm) and irrelevant frequencies (Assumption 2). We then divide the power in the irrelevant range with the total power. IPR can be used as an unsupervised measure of signal quality.

\subsection{Training}
\label{sec:method:training}
\textbf{Sampling.} When training, we randomly sample $W$ seconds from a video $X_i$ and its associated physiological ground truth $Y_i$. For our experiments, we set $W$ to ten seconds. We denote these subset clips as $x_a$ and $\tilde{y}_a$, respectively. We randomly augment our training sets by artificially stretching shorter video clips to $W$ seconds using trilinear interpolation. We use linear interpolation to mimic this stretching in the ground truth PPG and scale ground truth HR appropriately. At most, this effectively decreases the HR by 33\%. If the calculated IPR for a given sample PPG exceeds 60\%, we redraw a new $W$ second subset.

\textbf{Heart Rate Calculation.} Given a PPG, we calculate heart rate by (1) zero-padding the PPG signal for higher frequency precision, (2) calculating the PSD, and (3) locating the frequency with the maximum magnitude within relevant heart rate range. We use this method to both calculate missing HR ground truth and HR from predicted PPG. We chose a simple PSD-based method instead of a learned one to maintain determinism.

\textbf{Saliency Sampler.} If enabled, the input video $x_a$ is first passed through the saliency sampler ($S$). The resulting spatially warped video is denoted as $x_a^s$. The output of the saliency sampler can be used to verify the performance of the network, as explored in~\refSec{sec:exp:saliency}.

\textbf{Supervised Training.} When performing supervised training, only the top portion of~\refFig{system} is used. The input video clip $x_a^s$ is passed through the PPG Estimator $g_\theta$, producing the PPG estimate $y_a$. We then apply the selected supervised loss function between $\tilde{y}_a$ and $y_a$. 

\textbf{Contrastive Training.} When performing contrastive training, we randomly choose a resampling factor $R_f$ between 66\% and 80\%. We then pass the anchor video clip $x_a^s$ through the trilinear resampler $R$ to produce the negative sample $x_n^s$. This effectively increases the frequency of the heart rate by a factor of 1.25 to 1.5. Both $x_a^s$ and $x_n^s$ are passed through the PPG Estimator $g_\theta$, producing $y_a$ and $y_n$, respectively. We then resample $y_n$ using the inverse of $R_f$ to output the positive signal $y_p$, whose frequency should match $y_a$. Finally, we apply the contrastive loss function, MVTL, using the PSD MSE distance. For our experiments, we set the number of views ($V_{N}$) to four and the length ($V_{L}$) to five seconds. As this method is unsupervised, we also use the validation set for training.

\textbf{Further Training Details.} 
We implemented our models using PyTorch 1.7.0~\cite{paszke19pytorch} and trained each model on a single NVIDIA Tesla V100 GPU. We used a batch size of 4 for all experiments and set $w_s$ and $w_t$ to 1, unless otherwise stated. 
In all experiments we use the AdamW optimizer with a learning rate of $10^{-5}$ and train for a total of 100 epochs. After supervised training, we select the model from the epoch with the lowest validation loss. Because the contrastive training does not use labels, we instead choose the model with the lowest IPR on the training set.
\begin{table}
\begin{footnotesize}
\begin{center}
\begin{tabular}{lcrrc}
\toprule
Dataset & PPG & Subj. & Dur. (hrs) & Freely avail. \\
\midrule
\rowcolor{lightgray} COHFACE~\cite{heusch2017reproducible} & $\checkmark$ & 40 & 0.7 &  $\checkmark$ \\
ECG-Fitness~\cite{spetlik_visual_2018} &  & 17 & 1.7 &  $\checkmark$ \\
MAHNOB~\cite{soleymani2011multimodal} & \checkmark & 27 & 9.0 & \\
MMSE-HR~\cite{tulyakov2016self} & & 40 & 0.8 & \\
\rowcolor{lightgray} MR-NIRP-Car~\cite{nowara2020Driving} & $\checkmark$ & 19 & 3.3 &  $\checkmark$ \\
MR-NIRP-Indoor~\cite{magdalena2018sparseppg} & $\checkmark$ & 12 & 0.6 &  $\checkmark$ \\
OBF~\cite{yu2019remote} & $\checkmark$ & 106 & 177.0 &  \\
\rowcolor{lightgray} PURE~\cite{stricker2014non} & $\checkmark$ & 10 & 1.0 & $\checkmark$ \\
\rowcolor{lightgray} UBFC-rPPG~\cite{bobbia2019unsupervised} & $\checkmark$ & 42 & 0.8 & $\checkmark$ \\
VIPL-HR~\cite{niu_accv_2018_viplhr} & $\checkmark$ & 107 & 20.0 & \\
VIPL-HR-V2~\cite{Li_2020_CVPR_Workshops} & $\checkmark$ & 500 & 21.0 & \\
\bottomrule

\end{tabular}
\end{center}
\caption{\textbf{Survey of published datasets for rPPG analysis.} Not all datasets are available for open research. For our experiments, we selected the highlighted subset of RGB datasets containing a ground truth PPG signal. We did not use ECG-Fitness because it has extreme motion and utilizes ECG instead of PPG. We used MR-NIRP-Car in place of MR-NIRP-Indoor, as it was the more challenging set, containing body motion and lighting changes.\label{table:datasets}}
\end{footnotesize}
\end{table}

\section{Datasets}
\label{sec:datasets}

To test our model as fairly as possible, we evaluated it on four publicly available rPPG datasets from recent literature. Table~\ref{table:datasets} shows the datasets which we considered for evaluation. 
During the construction of this table, it became clear that much prior work had evaluated their methods using combinations of proprietary datasets and/or datasets which were not freely available to industrial researchers. This made replication of results expensive or impossible.
To avoid this, we opted to use freely available data for both training and evaluation. We ingested the following four datasets into a common data format, which we make available for other researchers to utilize. 

\textbf{PURE}~\cite{stricker2014non} consists of 10 subjects (8 male, 2 female) performing different, controlled head motions in front of a camera (steady, talking, slow translation, fast translation, small rotation, medium rotation) for one minute per sequence, under natural lighting. During these sequences, the uncompressed images of the head, as well as reference PPG and heart rate from a finger clip pulse oximeter were recorded. The first two samples of the PPG were corrupted and were discarded during analysis. PURE contains predefined folds for training, validation, and test, and we use these to be comparable to related work. We run each experiment 25 times with different random seeds and average our performance.

\begin{table*}
\begin{small}
\begin{center}
\begin{tabular}{l|rrr|rrr|rrr|rrr}

\toprule
 & \multicolumn{3}{c}{PURE} & \multicolumn{3}{c}{COHFACE} & \multicolumn{3}{c}{MR-NIRP-Car} & \multicolumn{3}{c}{UBFC} \\
\cmidrule(lr){2-4} \cmidrule(lr){5-7} \cmidrule(lr){8-10} \cmidrule(lr){11-13}
Method & RMSE & MAE & PC & RMSE & MAE & PC & RMSE & MAE & PC & RMSE & MAE & PC \\
\midrule
Mean & 16.0 & 13.0 & -0.27 & 15.4 & 12.0 & -0.19 & 15.1 & 12.7 & - & 21.8 & 19.1 & 0.00 \\
Median & 22.0 & 19.2 & -0.05 & 17.8 & 14.6 & -0.09 & 16.9 & 14.7 & - & 21.6 & 18.8 & -0.11 \\
HR-CNN
\cite{spetlik_visual_2018} & \textbf{2.4} & \textbf{1.8} & 0.98 & 10.8 & 8.1 & 0.29 & - & - & - & - & - & - \\
Nowara et al.~\cite{nowara2020Driving} & - & - & - & - & - & - & $^*$2.9 & - & - & - & - & - \\
Meta-rppg~\cite{lee_meta-rppg_2020} & - & - & - & - & - & - & - & - & - & $^*$7.4 & $^*$6.0 & $^*$0.53 \\
\midrule
Our Supervised & 2.6 & 2.1 & \textbf{0.99} & 7.8 & 2.5 & 0.75 & \textbf{1.6} & \textbf{0.7} & 0.96 & 4.9 & 3.7 & \textbf{0.95} \\
\hspace{0.08in}With Saliency & 2.6 & 2.1 & \textbf{0.99} & 7.6 & 2.3 & 0.76 & 1.8 & 0.8 & \textbf{0.97} & 5.0 & 3.8 & \textbf{0.95} \\
\midrule
Our Contrastive & 2.9 & 2.3 & \textbf{0.99} & \textbf{4.6} & \textbf{1.5} & \textbf{0.90} & 4.1 & 1.7 & 0.91 & \textbf{4.6} & \textbf{3.6} & \textbf{0.95} \\
\hspace{0.08in}With Saliency & 3.0 & 2.3 & \textbf{0.99} & 5.5 & 1.8 & 0.84 & 5.2 & 2.4 & 0.87 & 6.1 & 5.0 & 0.91 \\
\bottomrule

\end{tabular}
\end{center}
\caption{\textbf{Experiment Results.} Results on all datasets using our supervised and contrastive systems, with and without saliency, averaged over 25 independent training runs. We compare with mean and median baselines, as well as the strongest comparable baseline we could find for each dataset. Overall, our contrastive approach performs on par or better than existing methods for most reported results, and in some cases better than our own supervised training. $^*$Note that, for reasons described in~\refSec{sec:datasets} and \refSec{sec:experiments:dataset}, baselines for MR-NIRP-Car and UBFC are not directly comparable to our results. We make our full data pre-processing and evaluation pipeline available to support fair comparisons in the future. \label{table:mainresults}}

\end{small}
\end{table*}

\textbf{COHFACE}~\cite{heusch2017reproducible} consists of 160 one minute videos from 40 subjects, captured under studio and natural light and recorded with a Logitech HD Webcam C525 and contact rPPG sensor. The videos are heavily compressed using MPEG-4 Visual, which was noted by \cite{mcduff2017impact} to potentially cause corruption of the rPPG signal. Similarly to PURE, the dataset comes with preassigned folds and we run each experiment 25 times for stability.

\textbf{MR-NIRP-Car}~\cite{nowara2020Driving} is the first publicly available video dataset with ground truth pulse signals captured during driving. 
Data was captured simultaneously in RGB and near-infrared (NIR), with associated pulse oximeter recordings. The dataset contains 190 videos of 19 subjects captured during driving as well as inside a parked car in a garage. Each subject performed different motion tasks (looking around the car, looking at mirrors, talking, laughing, sitting still). To be consistent with~\cite{nowara2020Driving}, we only consider the subset of ``RGB garage recordings'', which have minimal head motion and consistent lighting. One sample had to be discarded due to a corrupted compressed file. Furthermore, we noticed that there were many stretches of zero values in the PPG signal that we had to resample around during training. Due to the lack of a set of folds, we split the dataset into five folds by subject id. We then conducted five training runs using a different held-out test set each time. We repeat this process five times, for a total of 25 runs per model, and then averaged the results across all runs.

\textbf{UBFC-rPPG}~\cite{bobbia2019unsupervised} contains uncompressed videos from 42 subjects, with ground truth PPG and heart rate data from a pulse oximeter. Heart rate variation was induced in participants by engaging them in a time-sensitive mathematical puzzle. To match the results of~\cite{lee_meta-rppg_2020}, for the purposes of testing we discard subjects 11, 18, 20, and 24, who were observed to have erroneous ($<5$ bpm) heart rate data. However, since only the PPG data and not the videos are corrupted, we include those videos for self-supervised training. Because no fold splits are included in the UBFC dataset, we use the same test strategy as with MR-NIRP-Car.

\textbf{Commonalities.} In all datasets, videos are captured at 30Hz. Where necessary, we interpolated the physiological data to synchronize it to get one sample for each video frame. Because we make the key assumption that heart rate does not vary over short time intervals (Assumption 2), we analyzed each of the datasets for the amount of variation in heart rate. In general, we found that heart rate did not vary by more than 2.5 bpm in the majority of data over a $10s$ window. Please see the supplementary material for further analysis.

\section{Experiments}
\label{sec:experiments}

\subsection{Dataset Performance}
\label{sec:experiments:dataset}

We first compare our method on a set of four recent PPG datasets in Table~\ref{table:mainresults}. As described in the previous section, while several larger datasets have recently been released~\cite{niu_accv_2018_viplhr,niu_video-based_2020}, we were unable to access them for benchmarking due to usage restrictions. We show the results of both our supervised and contrastive systems, with and without the use of a saliency sampler. We compare against two baselines which predict the mean and median of the test data. We calculate the root mean squared error (RMSE), mean absolute error (MAE), and Pearson's correlation (PC) of the predicted versus ground truth heart rate. We present further performance statistics over the PPG signals in the supplementary material.

\textbf{PURE.} The results across all systems, including the HR-CNN baseline~\cite{spetlik_visual_2018}, are similar for PURE - between 2.4 and 3.0 RMSE. This close-to-ideal performance is likely due to the high quality video, constrained environment, and minimal movement. Notably, the contrastive method is able to achieve similar results without the use of ground truth.

\textbf{COHFACE.} We find that our contrastive system performs best on COHFACE, with an RMSE of 4.6, despite not using labels during training. This provides evidence that our system is more robust to video compression versus other methods. While our supervised model performs worse (7.8 RMSE), it still outperforms the strongest comparable baseline (10.8 RMSE)~\cite{spetlik_visual_2018}. 

\textbf{MR-NIRP-Car.} To be comparable to~\cite{nowara2020Driving}, we report the RGB garage minimal motion subset. Although the baseline is already low at 2.9 RMSE, our supervised system improves this to 1.6 RMSE. Note that the baseline was knowledge-based and did not rely on training data, making the results not entirely comparable. Our contrastive system performs slightly worse (4.1 RMSE), but does so without ground truth labels.

\textbf{UBFC.} Lastly, we consider the UBFC dataset, comparing against the 7.4 RMSE baseline from~\cite{lee_meta-rppg_2020}. Note they instead use the first two seconds of all samples for adaptation, so the results are not perfectly comparable. Our supervised (4.9 RMSE) and contrastive (4.6 RMSE) methods both achieve a similar, improved performance.

\subsection{Saliency Sampler}
\label{sec:exp:saliency}
We evaluate our saliency sampler both quantitatively (whether or not it changes the PPG estimator's performance on the primary task of rPPG), and qualitatively (whether or not it aids in interpretation of the model's behavior).
Results from adding in the saliency sampler to the model when training for both supervised and contrastive models are shown in Table~\ref{table:mainresults}. 
In PURE, the sampler had no significant effect on performance, while for COHFACE, MR-NIRP-Car, and UBFC it alters performance by at most one point.
In~\refFig{SaliencyMaps} we show the qualitative effect of varying the sparsity and temporal regularization parameters during training. We include an experiment looking at the sensitivity to these parameters in the supplementary material.

We conjecture that while the temporal regularizer may particularly help performance in videos with little motion -- by allowing motion cues to pass through to the PPG estimator -- in videos with large motion it may hinder performance. On the other hand, the sparsity regularizer reliably helps to achieve intepretable saliency maps without harming performance significantly.

To illustrate the qualitative value of the sampler, we run a toy experiment injecting a periodic nuisance signal, as shown in~\refFig{Interpretability}. Under contrastive training, the sampler can be used to determine that the PPG estimator has found a spurious signal, and is not working as expected. Under supervised training, the sampler learns to remove the periodic signal from the input data to the PPG estimator altogether.

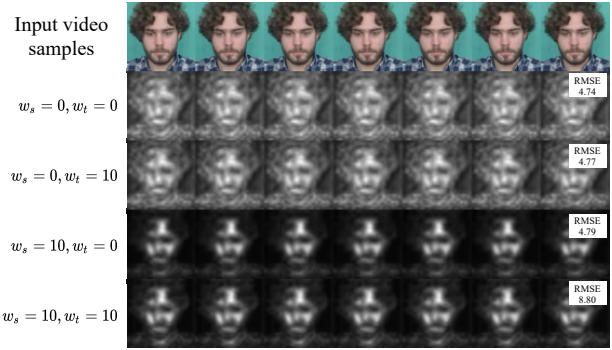
\myfigure{SaliencyMaps}{\textbf{Effect of regularization on the computed saliency estimates of a contrastively trained network.} The effect of the sparsity regularizer is especially clear. We note that the rPPG model performance was observed to be stable under various regularization settings, although training became empirically less stable at higher levels of regularization (as in the bottom row, caused by one of five training runs failing to converge).}

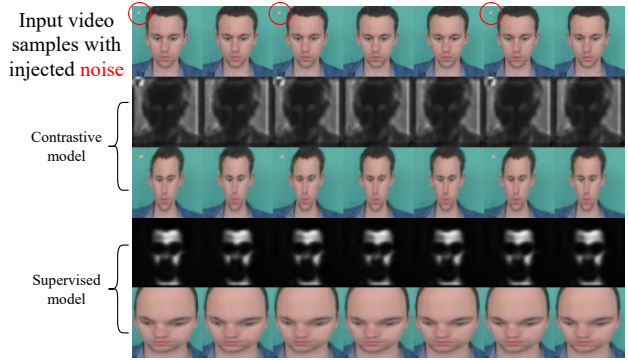
\myfigure{Interpretability}{\textbf{An example of interpretable model behavior.} We added a random flashing pixel block (highlighted in the top row) at 60-180 bpm to the UBFC dataset and trained both our contrastive model and our supervised model. In the contrastive case, without the need for ground truth validation data, the saliency map reveals that the model has learned to use the noise signal rather than learn the signal of interest on the subject. In contrast, in the supervised setting, the saliency sampler learns to emphasize the skin of the subject, and completely discards the injected noise signal after resampling, since it has no relevance to the prediction of PPG.}

\section{Discussion}
\label{sec:discussion}

In this paper we presented a contrastive approach to estimating cardiac activity in a person from video of their face. We believe this is the first time that deep neural networks have been applied to the problem of remote photoplethysmography in a fully self-supervised manner, at zero annotation cost. This would allow heart rate detection to be adapted to a specific domain without first acquiring labelled data in the domain.
We demonstrate the value of this approach through an accompanying workshop challenge submission~\cite{gideon21v4v}, in which self-supervised training on domain-shifted test data was shown to improve system performance.

In addition, we introduced a novel loss for supervised training that is more robust to ground truth synchronization error and yields improved performance. We also proposed the use of a saliency sampler to provide interpretable output to confirm whether the system is behaving as expected.

Our work opens the door to training on significantly larger, unlabelled datasets, such as sourced from the internet. This may help to improve the generalizability of heart rate estimation to more challenging domains and conditions, such as datasets with more severe lighting changes and head motion.
We also aim to further explore the use of learned saliency or attention-like mechanisms to more efficiently direct the efforts of a downstream PPG estimator, in a way that better conserves the original raw image pixels.

\vspace{3pt}
\noindent\textbf{Acknowledgments.}
We wish to thank Luke Fletcher for quickening our pulses at the right time, and the team at the Shapiro Cardiovascular Center at BWH from the bottom of (one of) our hearts.


\appendix

\section{Appendix}

\renewcommand{\thesection}{A\arabic{section}}

\section{Model Architecture}
\begin{table}[tp]
\begin{center}
\begin{footnotesize}
\begin{tabular}{lllllll}
\toprule
ENCODER & in & out & kernel & stride & pad\\
\midrule
Conv3d + BN3d + ELU & C & 32 & (1,5,5) & 1 & (0,2,2) \\
AvgPool3d & & & (1,2,2)& (1,2,2) & 0 \\
Conv3d + BN3d + ELU & 32& 64& (3,3,3) & 1 & (1,1,1) \\
\midrule
Conv3d + BN3d + ELU & 64& 64& (3,3,3) & 1 & (1,1,1) \\
AvgPool3d & & & (2,2,2)& (2,2,2) & 0 \\
Conv3d + BN3d + ELU & 64& 64& (3,3,3) & 1& (1,1,1) \\
\midrule
Conv3d + BN3d + ELU & 64& 64& (3,3,3) & 1 & (1,1,1) \\
AvgPool3d & & & (2,2,2)& (2,2,2) & 0 \\
Conv3d + BN3d + ELU & 64& 64& (3,3,3)& 1& (1,1,1) \\
\midrule
Conv3d + BN3d + ELU & 64& 64& (3,3,3)& 1& (1,1,1) \\
AvgPool3d & & & (1,2,2)& (1,2,2)&  0 \\
Conv3d + BN3d + ELU & 64 & 64 & (3,3,3)& 1& (1,1,1) \\        
\midrule
Conv3d + BN3d + ELU & 64 & 64 & (3,3,3)& 1& (1,1,1) \\
\bottomrule
\toprule
DECODER & \\
\midrule
Interpolate & & & (2,1,1) & & \\
Conv3d + BN3d + ELU & 64 & 64 & (3,1,1) & 1 & (1,0,0) \\
Interpolate & & & (2,1,1) & & \\
Conv3d + BN3d + ELU & 64 & 64 & (3,1,1) & 1 & (1,0,0) \\        
AdaptiveAvgPool3d & & & (-,1,1) & &\\
Conv3d & 64 & 1 & (1,1,1) & 1 & (0,0,0) \\
\bottomrule
\end{tabular}
\end{footnotesize}
\end{center}
\caption{\textbf{Modified PhysNet-3DCNN architecture}. The architecture follows an encoder-decoder structure with 3D convolutions to represent patterns through time; ``s'' corresponds to stride, ``p'' to padding, ``C'' to the number of input channels.}
\label{table:arch:physnet}
\end{table}
For the PPG estimator we use a modified 3D-CNN version of PhysNet~\cite{yu_remote_2019_physnet} as described in Table~\ref{table:arch:physnet}. Our modification is to use interpolation and convolution in the decoder instead of transposed convolution, which we found to reduce the aliasing artifacts that were present in the original model. 
For the saliency sampler, we use the architecture described in~\cite{recasens2018learning}, swapping out the saliency network for the shallower model shown in Table~\ref{table:arch:saliencynet} which was found to be sufficient for detecting facial parts (by the nature of the saliency maps learned).

\begin{table}[tp]
\begin{center}
\begin{footnotesize}
\begin{tabular}{lllllll}
\toprule
SALIENCY NET & in & out & kernel & stride & pad\\
\midrule
Conv2d + BN2d + ReLU & C & 64 & (1,7,7) & 2 & 3 \\
MaxPool & & & (1,3,3) & 2 & 1\\
BasicBlock & 64 & 64 & & 1 & \\
BasicBlock & 64 & 64 & & 1 & \\
BasicBlock & 64 & 64 & & 1 & \\
\bottomrule
\end{tabular}
\end{footnotesize}
\end{center}
\caption{\textbf{Saliency Network}. The architecture follows a Resnet-18 structure, truncated after \texttt{layer1}, with pre-trained ImageNet weights. Each BasicBlock consists of a $3\times3$ convolution, 2D batch normalization, ReLU, $3\times3$ convolution, 2D batch normalization, addition with the BasicBlock input (the residual) and a final ReLU. For further details see~\cite{he2016deep}.}
\label{table:arch:saliencynet}
\end{table}

\begin{table*}[tp]
\begin{small}
\begin{center}
\begin{tabular}{l|ccc|ccc|ccc|ccc}

\toprule
 & \multicolumn{3}{c}{PURE} & \multicolumn{3}{c}{COHFACE} & \multicolumn{3}{c}{MR-NIRP-Car} & \multicolumn{3}{c}{UBFC} \\
\cmidrule(lr){2-4} \cmidrule(lr){5-7} \cmidrule(lr){8-10} \cmidrule(lr){11-13}
Method & PC & MCC & SNR & PC & MCC & SNR & PC & MCC & SNR & PC & MCC & SNR \\
\midrule
Mean & -0.01 & 0.12 & -9.6 & -0.01 & 0.14 & -9.2 & -0.02 & 0.36 & 5.6 & 0.00 & 0.10 & -13.0 \\
Median & 0.00 & 0.08 & -10.9 & 0.00 & 0.14 & -9.2 & 0.00 & 0.30 & 1.4 & 0.01 & 0.10 & -13.5 \\
\midrule
Our Supervised & 0.54 & \textbf{0.90} & 18.1 & 0.23 & 0.57 & 15.4 & \textbf{0.52} & \textbf{0.79} & 17.8 & \textbf{0.17} & \textbf{0.64} & 13.3 \\
\hspace{0.08in}With Saliency & \textbf{0.58} & \textbf{0.90} & 18.1 & \textbf{0.30} & 0.57 & 15.5 & 0.42 & 0.78 & 17.9 & 0.15 & \textbf{0.64} & 13.3 \\
\midrule
Our Contrastive & 0.02 & 0.79 & 19.4 & -0.19 & \textbf{0.65} & 17.9 & 0.18 & 0.74 & \textbf{19.3} & 0.03 & 0.63 & \textbf{14.3} \\
\hspace{0.08in}With Saliency & 0.00 & 0.80 & \textbf{19.5} & -0.04 & \textbf{0.65} & \textbf{17.9} & 0.27 & 0.74 & 18.8 & 0.01 & 0.61 & 13.3 \\
\bottomrule

\end{tabular}
\end{center}
\caption{\textbf{Experiment PPG Statistics.} PPG statistics on all datasets using our supervised and contrastive systems, with and without saliency. We compare with both a mean and median baseline. The top performing system varies greatly depending on the dataset and statistic. However, the contrastive systems often perform comparable or better than the supervised ones when considering sync-robust metrics (without the need for ground truth). \label{table:ppgresults}}
\end{small}
\end{table*}

\section{Other Loss Functions/Metrics}
\textbf{Pearson's correlation} (PC) is commonly used as a loss and metric in other rPPG works (\eg \cite{yu_remote_2019_physnet}). While it is scale invariant, it assumes that there is perfect temporal synchronization between the ground truth and observed data. Otherwise the network must be capable of learning a temporal offset, assuming the offset is constant.

\textbf{Signal-to-noise ratio} (SNR) is another baseline used in prior rPPG works (\eg \cite{spetlik_visual_2018}) which train to match a ground truth heart rate instead of the full PPG signal. It relaxes the alignment assumption by expressing the loss in the frequency domain using the power spectral density (PSD). It calculates the amount of power in the target heart rate frequency bin of the PSD and compares it to the total other power in the PPG signal. Because of this, it assumes that all other frequencies should be zeroed out, which may remove meaningful harmonics.

\section{Dataset PPG Performance}
\label{sec:ppg_performance}

To supplement Table 3 from the main paper, we present further results on the four PPG datasets in Table \ref{table:ppgresults}, using metrics which capture statistics of predicted vs.~ground truth PPG signals (as opposed to the final predicted heart rate). We again show the results of both our supervised and contrastive systems, with and without the use of a saliency sampler. However, as these PPG statistics are not given for the other cited baseline systems, we instead provide only the mean and median methods as baselines.

We calculate the Pearson's correlation (PC), Max cross-correlation (MCC), and Signal-to-noise ratio (SNR) of the predicted versus ground truth PPG signals. While we present PC for comparison, we expect MCC and SNR to be better measures of system performance, as they are calculated in the frequency domain. This makes them more robust to desynchronization between predictions and ground truth. Supervised training uses MCC as a loss function and validation metric, while contrastive training works without guiding ground truth. Because of this, it is possible for either system to learn a random phase offset, as long as the overall frequency information is predictive of heart rate.

\textbf{PC.} Across all dataset results, we note that supervised training attains the highest PC. Even though desynchronization is not penalized during training, the model likely learns the easiest mapping between video and ground truth - one without an additional offset. This could indicate that the datasets only have minimal offset between observation and ground truth.

\textbf{MCC.} We also note that supervised training tends to result in higher MCC, which is likely due to the guiding ground truth. Without ground truth, the contrastive method is only able to learn the periodic signal visible in the input video. The supervised method would be able to learn to replicate any repeating artifacts of the biometric PPG sensor, producing a stronger MCC. However, the overall performance on the contrastive COHFACE model is substantially better than that of the supervised one, as seen in the heart rate results. This likely lessens the relative impact of PPG artifacts, when compared with other datasets with closer performance.

\textbf{SNR.} Unlike MCC, SNR penalizes the learning of all other frequencies besides heart rate. So perfect PPG prediction can result in lower SNR performance, if the PPG includes other frequencies. Because the supervised training uses MCC as a loss function and the ground truth isn't a perfect sine wave, this encourages sub-optimal SNR. We see this reflected in the results, with contrastive learning having a higher SNR across all datasets.

\section{Loss Function Robustness}

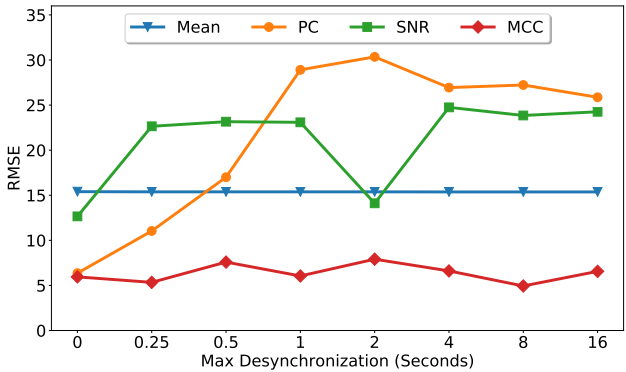
\myfigure{loss_robustness}{\textbf{Loss Function Robustness to Desynchronized Ground Truth.} The performance of supervised training on COHFACE with varying amounts of random desynchronization applied between the video and the ground truth PPG signal. We show the performance of 3 loss functions (PC, SNR, MCC) against a Mean model baseline (which estimates the mean heart rate in the test set). Unlike other loss functions, MCC is shown to be robust to ground truth desynchronization. \vspace{-3mm}}

In Section~\ref{sec:ppg_performance}, we examined how the metrics PC, MCC, and SNR can be used to gauge the performance of PPG prediction versus ground truth. In this section, we examine how each metric performs as a loss function during supervised training, as well as the impact of desynchronization between the observed video and the ground truth PPG.

We use the COHFACE dataset and compare versus a baseline that always predicts the mean heart rate in the test set. We select the maximum amount of injected synchronization error ($O_{\textnormal{max}}$) to range between 0 and 16 seconds. Each time a new clip is drawn during training, a random offset is chosen between $-O_{\textnormal{max}}$ and $O_{\textnormal{max}}$ using uniform sampling. We then shift the ground truth PPG by the selected offset using neighboring data. We train a model using the supervised pipeline and the selected loss function. We calculate the selected loss on held-out validation data each epoch and use the model with the lowest loss at test time. In this experiment, we do not use the saliency sampler since the purpose is to explore the robustness of supervised loss functions.

Figure~\ref{fig:loss_robustness} shows the RMSE performance of our supervised system for different $O_{\textnormal{max}}$ and loss functions. Without injected desynchronization, we find that PC and MCC perform similarly. This likely indicates that ground truth in COHFACE is consistently aligned with the video, either with a minimal or constant (learnable) offset. Because MCC is the offset-adjusted version of PC, we note that they have similar performance without the presence of offsets. However, when increasing amounts of synchronization error are applied, the performance of the system trained with PC quickly degrades, while the one trained with MCC remains relatively stable. We also note that SNR has consistently poor performance at all offsets, indicating that it is a weaker supervisory signal for rPPG compared to correlation-based measures. Based on these results, we favored MCC as a loss function for supervised training, particularly if synchronization issues were suspected to be present in the training data.

\section{Dataset Statistics}
Our approach relies on the assumption (Assumption 2 in the main paper) that ``the signal of interest typically does not vary rapidly over short time intervals: the heart rate of a
person at time $t$ is similar to their heart rate at $t + W$, where $W$ is in the order of seconds.''
In~\refFig{HR_diff_per_time_offset}, we show the distribution of heart rate variation at different time intervals within each of the four datasets.
With approximately 80\%-100\% certainty, depending on the dataset, one can assume that the heart rate at $t+10$s is within $10$bpm of the heart rate at time $t$. All datasets consistently have a median variation of about $2.5$bpm over a $10$s period. UBFC was found to contain the most heart rate variability of the datasets examined.

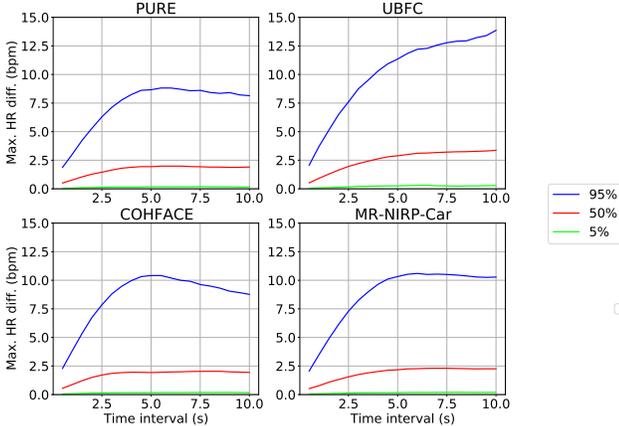
\myfigure{HR_diff_per_time_offset}{\textbf{Distribution of heart rate differences at different time intervals apart}. The 5\%, 50\% and 95\% quantile lines are shown.}

\begin{table*}
\begin{footnotesize}
\begin{center}
\begin{tabular}{r|rrrr|rrrr|rrrr}
\multicolumn{1}{c}{} & \multicolumn{4}{c}{RMSE change ($\downarrow$)} & \multicolumn{4}{c}{MAE change ($\downarrow$)} & \multicolumn{4}{c}{PC change ($\uparrow$)} \\ 
\midrule
$w_s$ \textbackslash \; $w_t$ & 0 & 0.1 & 1 & 10 & 0 & 0.1 & 1 & 10 & 0 & 0.1 & 1 & 10 \\
\midrule
0 & - & 1.4 & 1.4 & -0.9 & - & 0.7 & 0.6 & -0.2 & - & -0.06 & -0.06 & 0.02 \\
0.1 & 3.1 & 0.8 & -0.8 & 0.1 & 1.3 & 0.4 & -0.2 & 0.0 & -0.18 & -0.04 & 0.02 & 0.00 \\
1 & -0.8 & 0.0 & -0.7 & -0.4 & -0.2 & 0.0 & -0.2 & -0.1 & 0.02 & 0.00 & 0.02 & 0.01 \\
10 & 0.2 & 0.6 & 0.4 & -0.9 & 0.0 & 0.2 & 0.2 & -0.2 & 0.00 & -0.02 & -0.01 & 0.03 \\
\bottomrule
\end{tabular}
\end{center}
\caption{\textbf{Sensitivity of saliency sampler regularization on rPPG performance for a contrastively trained model on UBFC.} As with Table 3 from the main paper, we average the results of five runs on different folds with held-out subjects, and report the performance differences relative to the zero regularization case, ($w_s$, $w_t$) = (0,0). rPPG performance tends to improve when the regularization parameters are set in the range {[1,10]}, although the best parameters depend on variables such as dataset and resolution.\label{table:ss-reg-ubfc}}
\end{footnotesize}
\end{table*}

\section{Unsupervised Learning Protocol}
In our contrastive experiments, since our method does not utilize ground truth labels, we fold validation data into the training set. In other words, in Table 3 of the main paper, the contrastive models ``see'' a little more training data than the supervised models.
When validation data is excluded from contrastive training, we found test RMSE to worsen by 0.7 bpm on average across datasets. This shows the value of larger training sets when training with contrastive loss. Exploring the trade off between training data size and model performance is a topic for future work.

\section{Additional Baseline}
While we selected the strongest comparable baseline we could find for each dataset, one reviewer requested the addition of the Siamese CNN proposed by~\cite{tsou2020siamese}.
However, we note that this work is not directly comparable for several reasons, in particular: (i) for UBFC evaluation, we do not pre-train our model on COHFACE; (ii) the final results reported in~\cite{tsou2020siamese} use a 20s (COHFACE) and 30s (UBFC, PURE) input time window, while we use a shorter 10s window consistently across datasets. In Table 2 of~\cite{tsou2020siamese} the effect of window length is shown for COHFACE: they report 1.8 RMSE with window length 400, and 4.7 with 256. Our \textit{self-supervised} baseline achieved 4.6 RMSE with window length 300 (25 run average). This shows that over a range of datasets we can achieve performance comparable to supervised deep learning methods (\textit{e.g.}~\cite{tsou2020siamese}) using our approach \textit{without annotations}.

\section{Sensitivity to Regularization Parameters}
\label{sec:reg-sensitivity}

In Table~\ref{table:ss-reg-ubfc} we show the relative performance of a contrastive model trained on the UBFC dataset as the saliency sparsity regularization weight, $w_s$, and temporal regularization weight, $w_t$, are varied in the range {[0, 0.1, 1, 10]}. We find that model performance is not significantly impacted. 
The saliency map output is most visible when higher values of the sparsity term $w_s$ are used, although we observe that the best parameters can depend on variables such as dataset and resolution.

{\small
\bibliographystyle{ieee_fullname}
\bibliography{references}
}

\end{document}